\documentclass{SCIS2018}
\begin{document}

\ArticleType{LETTER}
\Year{2018}
\Month{}
\Vol{61}
\No{}
\DOI{}
\ArtNo{}
\ReceiveDate{}
\ReviseDate{}
\AcceptDate{}
\OnlineDate{}

\title{Incremental Data-driven Optimization of Complex Systems in Nonstationary Environments}{Title for citation}

\author[1]{Cuie YANG}{}
\author[1]{Jinliang DING}{{jlding@mail.neu.edu.cn}}
\author[1,2]{Yaochu JIN}{{yaochu.jin@surrey.ac.uk}}
\author[1]{Tianyou CHAI}{}
\AuthorMark{Author A}

\AuthorCitation{Author A, Author B, Author C, et al}


\address[1]{State Key Laboratory of Synthetical Automation for Process Industries, Northeastern University, Shenyang, 110004, China}
\address[2]{Department of Computer Science, University of Surrey, Guildford, Surrey GU27 XH, U.K.}
\maketitle

\begin{multicols}{2}
\deareditor

In a complex system, many real-world optimization problems do not have explicit optimization functions or constraint functions, whereas only the data from production processes is available due to the complexity of the system~\cite{1}. Thus the optimization in this scenario can generally rely on the collected historical datasets and these problems are also known as offline data-driven optimization problems~\cite{2}. In the last decade, the literature for offline data-driven optimization works by first constructing a surrogate model with the collected data and then considering the best solution of the surrogate model as the optimal decision~\cite{1}, little work has been considered the errors existed in surrogate models. In addition, production processes often subject to changes, leading to the attainment of a data that are non-independent identically distributed of the obtained data. In this letter, we aim to deal with the offline data-driven optimization problem that the data generated in nonstationary systems in an incremental manner, specifically, we consider the data comes in a chunk as shown in in Figure 1 in the supplementary section. This type of data-driven optimization poses new challenges to the current data-driven optimization algorithms~\cite{2}, in which we focus on the following three issues. The first is how to build a high quality surrogate model for each environment. The second is about the optimization, that is, how to quickly exploit optimal solution of each new environment. The last is final solution creation, this is because the errors existed between surrogate models and corresponding real fitness functions (the unknown formulations of real systems), resulting in a less reliability of the solutions from surrogate models. In order to alleviate the above difficulties, this letter suggests a general method as described in follows:
\\ \textbf{Step 1:} \textbf{WHILE} new data chunk ${D_t}$ is income from complex systems at the $t-th$ environment
\\ \textbf{Step 2:}Update the surrogate model based on knowledge transfer technique to adapt the $t-th$ environment;
\\ \textbf{Step 3:}Initialize population based on historical knowledge;
\\ \textbf{Step 4:}Optimize the surrogate model by using DE algorithm;
\\ \textbf{Step 5:}Produce the final solution for complex systems;
\\ \textbf{Step 6:} \textbf{END WHILE}

\lettersection{Approach} A general framework of the proposed approach is listed as above. At the observation of each new data chunk from complex systems, the surrogate model is first updated to adapt to the new environment. A differential evolution (DE) algorithm~\cite{7}, a kind of population based global optimization approach, is adopted as the optimizer. Thus, the next step is a population initialization for the DE algorithm. Then the DE algorithm with the rand/1 strategy is applied to optimize the surrogate model in order to obtain optimal landscape of real fitness problems. Lastly, the final solution is generated from the searched superior landscape. Details of surrogate model update, population initialization and final solution production are presented as follows.

\begin{itemize}

\item Knowledge transfer based surrogate model adaptation: Ensemble approaches are a popular method to handle incremental learning, which uses models or training instances of historical environments to improve model quality of the current environment~\cite{5}. Nevertheless, most work in incremental learning focuses on classification tasks~\cite{5}, in the optimization problem of this letter, we introduce an ensemble approach for regression tasks to formulate the surrogate model. To be specific, a base regression learner is first trained via the new data chunk ${D_t}$, equivalently $({{\bf{x}}_t},{{\bf{y}}_t})$, and denoted as ${h_t}$. The radial basis neural network (RBFN) is applied as a base learner in this work due to its universal approximation ability~\cite{6}. After that, a set of base learners trained by using data chunks of each past environment is also constructed separately. To improve the adaptability of past training instances, we first map the historical data chunks ${D_1},{D_2},...,{D_{t - 1}}$ to the current data chunk space ${D_t}$, thus to facilitate the knowledge transfer between the historical data set and the current data set. Then use the combination of each of the transferred historical data chunk ${D_{ni}}$ and the current data chunk ${D_t}$ to build each historical base surrogate model ${h_i}$. Note that, we are interested in dynamics of the system, which would result in the change of function the value ${{\bf{y}}_t}$. Thus, we transform ${{\bf{y}}_i},i = 1,2,...,t - 1$ in each ${D_i}$ to the current objective space ${{\bf{y}}_t}$ by using Eq.~\ref{eq1}. we denote the transferred ${D_i}$, which contains $({{\bf{x}}_i},{{\bf{y}}_{ni}})$, as ${D_{ni}}$, $i = 1,2,...,t - 1$.

\begin{equation}\label{eq1}
{{\bf{y}}_{ni}} = \frac{{{{\bf{y}}_{i}} - y^{min}_i}}{y^{max}_i} - {y^{min}_i} \times ({y^{max}_t} - {y^{min}_t}) + {y^{min}_t}
\end{equation}
where ${y^{max}_i}$ and ${y^{min}_i}$ are the maximum and minimum value of ${{\bf{y}}_i}$, ${y^{max}_t}$ and ${y^{min}_t}$ are the maximum and minimum value of ${{\bf{y}}_t}$, respectively.

\end{itemize}

In the next step, all the base surrogate model ${h_i}$, $i = 1,2,...,t$ are integrate into final perfect surrogate $f$ by using Eq.~\ref{eq2},

\begin{equation}\label{eq2}
f = \frac{{\sum\limits_{i = 1}^t {{w_i}{h_i}} }}{{\sum\limits_{i = 1}^t {{w_i}} }}
\end{equation}
where ${w_i} = \frac{1}{{RMS{E_i} + RMS{E_t}}},i = 1,2,...,t - 1$ , ${w_t} = \frac{1}{{RMS{E_t}}}$ and $RMS{E_i} = \sqrt {\frac{1}{{\left| {{D_i}} \right|}}\sum\limits_{y \in {{\bf{y}}_i}}^{} {{{(\hat y - y)}^2}} } ,i = 1,2,...,t$.

In this ensemble, we assign a larger ${w_t}$ than ${w_i}, i = 1,2,...,t$ to ensure a higher weight of the base model in the current environment. The reason is that the data set of the current environment is more reliable, thus it should be fully used.
\begin{itemize}
\item A priori knowledge based population initialization: After the surrogate model of the current environment is obtained, an initial population should be created before the surrogate model optimization is started. The initial population is often randomly generated in the decision space in traditional DE algorithms. In reality, the surrogate models of different environments are not isolate since their training instances are from the same system. Therefore, using the historical knowledge of the past environments in the population initialization would benefit convergence of the current environment. For simplicity, the candidates of the latest environment are applied as the initial population in this work. Note that, the initial population is randomly generated for the first environment because there is no historical information in the beginning.

\item Best solution averaging based final solution production: As mentioned above, solutions obtained during optimization are not allowed to be evaluated by true complex problems, instead they are only evaluated by surrogate models. In this case, it is interesting to create a high quality final solution for real fitness functions since no surrogate can be updated using the real fitness function and the fitness value of a solution evaluated by surrogates may exist large errors with that evaluated by real fitness problems. This letter proposes a best solution averaging technique to generate final solution for real fitness function instead of directly using the best solution of the obtained candidates. Specifically, in the final population of each environment, the average of the top 10 percent best individuals is considered as the final solution. In this way, the errors of the final solution induced by surrogates can be smoothed by consulting a number of candidates.

\end{itemize}

\lettersection{Experimental results} The six dynamic optimization benchmark problems~\cite{8} are applied to examine the transferred surrogate model construction, population initialization and final solution production strategies. The number of decision variables $D$ of each problem is set to 10. The total number of environments in each test problem is set to 50. In each environment, $3D$ points generated by Laplace sampling and evaluated by the real fitness function are taken as historical dataset, where $D$ is the number of decision variables. The experiment conducts on different approaches of incremental data-driven optimization in non-stationary environments, SS (Single dataset based Surrogate model construction techniuqe), KTS (Knowledge Transfer based Surrogate model construction), KTSPI (the version of KTS by inducing a Prior knowledge based population Initialization), KTTLSA-TBA (the KTSPI algorithm with Best solution Averaging based final solution production technique) to verify each of the proposed technique.

\begin{table}[H]
\scriptsize
\caption{The average result of 50 environments over 20 independent runs}
\label{tab1}
\begin{center}
\tabcolsep 4pt 
\begin{tabular}{c c c c c}
\hline
Name & SS & KTS & KTSPI & KTSPI-TBA \\ \hline
F1 & 3.7472 & 3.4762 & 3.0220 & 2.9770 \\ \hline
F2 & 578.2938 & 546.8831 & 528.7354 & 528.9487 \\ \hline
F3 & 1.121e+03 & 1.094e+03 & 1.077e+03 & 1.074e+03 \\ \hline
F4 & 646.0489 & 610.8318 & 592.6871 & 589.7063 \\ \hline
F5 & 2.021e+03 & 2.016e+03 & 1.999e+03 & 1.997e+03 \\ \hline
F6 & 2.341e+03 & 1.399e+03 & 1.317e+03 & 1.314e+03 \\ \hline
\end{tabular}
\end{center}
\end{table}

Table~\ref{tab1} presents the average results of 50 environment over 20 independent runs of the compared algorithms. We can see from this table that, the value that obtained by SS, KTS, KTSPI, KTSPI-TBA becomes smaller in general, indicating the suitability of the knowledge transfer based surrogate model update technique, historical knowledge population initialization technique and averaging based final solution production technique. The result on each environment over 20 independent runs of the compared algorithms on F1 and F5 is presented in Figure 2 in the supplementary section. This figure shows an obvious superior performance of KTSPI and KTSPI-TBA in comparison with SS and KTS on most of the experiments on F1. On the F5 problem, we can find that KTS clearly outperforms SS on about all experiments, in addition, it also can be seen that the KTSPI-TBA algorithm achieves a more robust performance along the experiment compared with KTSPI.

\lettersection{Conclusion} This letter proposes a general method to solve data-driven problem in nonstationary environment which includes three techniques, i.e., knowledge transfer based surrogate model adaptation technique, historical knowledge based population initialization technique and the best solution averaging based final solution production technique . We systematically compare each technique by applying the four proposed incremental data-driven optimization approaches on six benchmark problems, the statistical results reveal the promising performance of each strategy in addressing the incremental offline data-driven problem in dynamic environments.

\Acknowledgements{This work was supported by the National Natural Science Foundation of
China under Grand 61525302, 61590922, the Project of Ministry of Industry
and Information Technology of China under Grand 20171122-6, the Projects
of Shenyang under Grand Y17-0-004, the Fundamental Research Funds for
the Central Universities under Grand N160801001, N161608001, and the
Outstanding Student Research Innovation Project of Northeastern University
under Grand N170806003.}



\end{multicols}
\end{document}